# IMPACT OF EXPLAINABLE AI ON COGNITIVE LOAD: INSIGHTS FROM AN EMPIRICAL STUDY

*Research Paper*

Lukas-Valentin Herm, University of Würzburg, Würzburg, Germany
lukas-valentin.herm@uni-wuerzburg.de

## Abstract

*While the emerging research field of explainable artificial intelligence (XAI) claims to address the lack of explainability in high-performance machine learning models, in practice, XAI targets developers rather than actual end-users. Unsurprisingly, end-users are often unwilling to use XAI-based decision support systems. Similarly, there is limited interdisciplinary research on end-users' behavior during XAI explanations usage, rendering it unknown how explanations may impact cognitive load and further affect end-user performance. Therefore, we conducted an empirical study with 271 prospective physicians, measuring their cognitive load, task performance, and task time for distinct implementation-independent XAI explanation types using a COVID-19 use case. We found that these explanation types strongly influence end-users' cognitive load, task performance, and task time. Further, we contextualized a mental efficiency metric, ranking local XAI explanation types best, to provide recommendations for future applications and implications for sociotechnical XAI research.*

## 1 Introduction

Due to recent advances in computing, the spectrum of potential use cases for the application of artificial intelligence (AI) is constantly expanding, enabling end-users to rely almost solely on data-driven decision support systems (DSS) (Berente et al., 2021, Janiesch et al., 2021). That is, integrating AI into information systems forms intelligent systems to enhance end-users' and organizations' effectiveness (Gregor and Benbasat, 1999, Herm et al., 2022). In this context, AI refers to an abstract concept mimicking human cognitive abilities through the application of mathematical and statistical algorithms, to generate (i.a.) machine learning (ML) models capable of automatically finding nonlinear relationships within data. So, decision knowledge is derived without the need for explicit programming (Goodfellow et al., 2016, Russell and Norvig, 2021). Research has focused on overcoming algorithmic constraints by increasing the decision complexity of ML algorithms, resulting in ML applications capable of outperforming domain experts even in complex and high-stakes use cases (Janiesch et al., 2021, McKinney et al., 2020). Furthermore, a subclass of ML algorithms, called deep learning (DL), uses deep neural network architectures to achieve unsurpassed performance. In turn, the inner decision logic of these models is no longer traceable by humans, which reduces end-users' willingness to use these AI-based DSSs; thus, their overall acceptance is decreased, potentially leading to algorithm aversion (Berger et al., 2021, Wanner et al., 2022).

To address this issue, the research stream of explainable AI (XAI) has developed approaches to overcome the lack of traceability while maintaining the performance of these black-box models (Meske et al., 2022). However, as all that glitters is not gold, these approaches are mostly mathematically driven models that provide technical explanations, as opposed to addressing the actual end-users of the system with a sound explanatory scope. That is, recent XAI approaches have mainly been designed by developers for developers (Arrieta et al., 2020, van der Waa et al., 2021). Consequently, first research endeavors emerged proposing research agendas (Laato et al., 2022), first-hand end-user evaluations (Herm et al., 2023, Shin, 2021), and design knowledge (Herm et al., 2022, Meske and Bunde, 2022). Yet, it is not completely apparent how an end-user's heuristic mental model behaves, in terms of different perception factors, when operating within a use-case (Laato et al., 2022). Unsurprisingly, IS research calls for further examinations of AI-based explanations from a sociotechnical perspective (Gregor and Benbasat, 1999, Herm et al., 2023), which also interfere with the research streams of human-computer interaction and cognitive science (Langer et al., 2021, Liao and Varshney, 2022).

Crucially, an explanation is a social and cognitive process of knowledge transfer from an XAI-based DSS to the end-user (Miller, 2019). It is unclear how end-users perceive these explanations, as increased cognitive load may be imposed when end-users rely on them to solve real-world tasks (Hudon et al., 2021). Furthermore, it is unknown whether this increased cognitive load affects end-user performance or the time required to solve a task (Hemmer et al., 2021). Consequently, XAI explanations should be perceived as mentally efficient to prevent end-users from feeling overwhelmed, stressed, and unable to perform well (Buçinca et al., 2020, Paas et al., 2016). Complicating matters further, due to the increasing attention to XAI in research and practice, numerous XAI applications are being developed, creating an XAI jungle from which to select an appropriate XAI approach (Das and Rad, 2020, Dwivedi et al., 2022). That is, organizations must determine how XAI explanations affect end-user behavior and what type of explanation should be used to form a sound DSS application within intelligent systems (Gregor and Benbasat, 1999). Hence, Mohseni et al. (2021) proposed an initial systematization that groups XAI explanation types in an implementation-independent manner, providing a foundation for future research and further facilitating generalizable findings that can be transferred to any type of XAI application in practice.

Following Mohseni et al. (2021)'s systematization, we contribute to IS research (Gregor and Benbasat, 1999, Meske and Bunde, 2022) by comparing these explanation types in an end-user-centered manner. Therefore, we conduct an empirical study in the field of medicine using COVID-19 X-ray images to measure end-users' cognitive load. Similarly, we benchmark end-users' task performance and time required to solve the task. Lastly, we combine these findings to put the mental efficiency metric of Paas et al. (2016) into the context of sociotechnical XAI research. To summarize our research intent, we propose the following research question (RQ):

> **RQ:** *Do XAI explanation types affect end-users' cognitive load and what are the ramifications for task performance and task time?*

The remainder of this paper is organized as follows: Section 2 presents the theoretical foundations, related work, and our measurement model. Section 3 describes the research methodology by following Müller et al. (2016) and the applied study design. Section 4 presents the data analysis, including demographic data, descriptive statistics, and hypotheses testing. Then, Section 5 discusses the findings to answer our RQ, derives implications for research and practice, and describes the study's limitations and recommendations for future research. Finally, Section 6 summarizes our research findings by drawing conclusions.

## 2  Theoretical Foundation

### 2.1  (Explainable) Artificial Intelligence

**Artificial Intelligence.** Following Berente et al. (2021), AI can be envisioned as an arbitrary frontier of computational advancements that mimics human-like or superhuman intelligence, enabling DSSs to assist end-users in accomplishing any task. A DSS employs ML models to enable these artificial cognitive capabilities. Here, ML is an umbrella term that encompasses mathematical and statistical algorithms used to automatically infer decision knowledge using historical data (Goodfellow et al., 2016). To this end, recent research has developed increasingly complex algorithms with high predictive power, making the models' rationale less tractable (Janiesch et al., 2021). Unsurprisingly, research has derived the performance-explainability trade-off, where inherently understandable models have been proposed to have the lowest performance and – conversely – DL models to have the highest performance (Herm et al., 2023). Here, DL is subsumed under the umbrella term ML and refers to a deep neural network architecture with decision logic that is no longer comprehensible to humans (Janiesch et al., 2021). DL applications can generate promising outcomes, even in high-stakes use cases (e.g., medicine) where a wrong decision could cost human lives (Dwivedi et al., 2022, McKinney et al., 2020). However, this may reduce end-users' willingness to use the system as non-traceability could lead to ambiguity and uncertainty in task solving (Epley et al., 2007). Alternatively, end-users may not be allowed to use the system due to regulations, such as the General Data Protection Regulation (GDPR) (Goodman and Flaxman, 2017).

**Explainable Artificial Intelligence.** In response, the multidisciplinary research stream of XAI has emerged. Its objective is to develop transfer techniques that make these opaque black-boxes comprehensible to users while preserving the predictive power of the underlying DL model (Arrieta et al., 2020, Meske et al., 2022). Thus, post-hoc explainability methods have been developed for specific types of ML models (model-specific) or a subset of them (model-agnostic); for different dataset formats (e.g., images, text, or tabular); and for different task types (e.g., classification or regression). They can also be distinguished by the nature of their explanatory scope – either explaining predictions for individual observations (local) or explaining the ML model's inner decision logic (global) (Das and Rad, 2020, Speith, 2022). This results in a plethora of explanation possibilities for depicting a rationale. In conjunction, countless distinct XAI applications have been developed in practice, creating an XAI jungle from which to choose and thus complicating the development process. Therefore, Mohseni et al. (2021) systematized these explanation types in an application-independent fashion. Table 1 summarizes these explanation types, their respective descriptions, and an exemplary excerpt of XAI's implementation jungle for each explanation type:

| Type[1] | Description[1] | Exemplary Implementations[2] |
|---|---|---|
| *How* | Holistic representation of how the ML model's inner decision logic operates – global explanation type. | ProfWeight, SHAP, DALEX, Saliency |
| *How-To* | Hypothetical adjustment of the ML model's input yielding a different output (counterfactual explanation) – local explanation type. | DiCE, KNIME, PDP |
| *What-Else* | Representation of similar instances of inputs that result in similar ML model outputs (explanation by example) – global explanation type. | SMILY, Alibi |
| *Why* | Description of why a prediction was made by informing which input features are relevant to the ML model – local explanation type. | SHAP, LIME, ELI5, Anchor |
| *Why-Not* | Description of why an input was not predicted to be a specific output (contrastive explanations) – local explanation type. | CEM, ProtoDash |

Legend: *1)* Types and definitions adapted from Mohseni et al. (2021); *2)* exemplary classification of frequently mentioned XAI implementation packages based on Das and Rad (2020), Dwivedi et al. (2022), Liao and Varshney (2022), and Mohseni et al. (2021).

Table 1. *Description and Implementation of XAI Explanation Types*

In practice, this XAI jungle is exacerbated by developers primarily designing these XAI implementations for developers without prioritizing the actual end-users (van der Waa et al., 2021). As first research endeavors target the end-user of an XAI-based DSS, these interdisciplinary research outcomes must be incorporated into practical applications to design valuable explanations for end-user (Arrieta et al., 2020). Following the explanation theory of Miller (2019), a useful explanation is defined as a social and cognitive process of knowledge transfer from an XAI-based DSS to the end-user. Thus, if an explanation is perceived as inadequate, contradicts an end-user's mental model, or does not appeal to their emotions or beliefs, trust issues can occur and user acceptance may be reduced, leading to algorithm aversion (Berger et al., 2021, Shin, 2021). Following recent IS research, a mental model defines any type of mental representation used to encode beliefs, facts, and knowledge when conceptualizing cognitive processes (Bauer et al., 2023). In this sense, the extent to which these explanation types affect end-users' cognitive load is unknown, which is an essential factor in the design and development of appropriate XAI implementations (Herm et al., 2023).

## 2.2 Cognitive Load Measurement

Although the human cognitive system can be considered an information-processing engine, its capacity is limited when using information systems. Providing too much or distracting information in an instructional design can lead to a high cognitive load for the end-user (Bahari, 2023). The cognitive fit theory (CFT) (Vessey, 1991) posits the relationship between a task and the required information presentation (i.e., the type of XAI explanation), where an inappropriate explanation type leads to poor end-user task performance. Moreover, end-users are unlikely to have a solid understanding of the instructional design or to build a representative mental model of the task problem (Simon, 1955). This leads to them feeling overburdened, stressed, and incapable of performing sound decision-making (Anderson et al., 2020, Paas et al., 2004). Therefore, cognitive research developed a computational approach that combines mental effort, task performance, and task time into a quantitative variable called mental efficiency to classify the goodness of instructional design with respect to end-users' information processing to prevent excessive mental workload in complex cognitive tasks (Paas et al., 2016). Accordingly, XAI explanations should require an appropriate level of cognitive load to represent the model's decision and facilitate seamless knowledge transfer to the end-user. Paired with an appropriate level of task performance and task time, the high mental efficiency of an XAI explanation constitutes a well-designed XAI-based DSS (Herm et al., 2023, Hudon et al., 2021).

While cognitive load is a multifaceted construct comprising various components, cognitive science research has developed several approaches for measuring it. Objective measures exist, such as eye activity, along with subjective measures, such as self-reported mental effort (Schmeck et al., 2015). While the former focuses on the identification of unconscious factors among participants, the latter targets conscious factors. Accordingly, both approaches behave complementary (Tams et al., 2014).

## 2.3 Preliminaries and Research Gap

To investigate the extent to which cognitive load from the perspective of XAI explanations has already been researched, we conducted a structured literature review according to Webster and Watson (2002). We focused on the information systems-related databases ScienceDirect, AIS eLibrary, and Web of Science, as well as the computer science-related databases ACM Digital Library and IEEE Xplore. Specifically, we used the following search term: *"((expla\* | interpreta\*) AND (explainable artificial intelligence | artificial intelligence | deep learning | machine learning | AI | XAI)) AND (cognitive load | mental load | mental effort | mental workload | cognitive capacity)"*. Without restricting our search in terms of (journal) rankings, we identified $n = 2,814$ publications as potentially relevant. Hence, we consider publications that examine or discuss the effects of XAI explanations (packages) on end-users' cognitive load as relevant. This results in $n = 17$ publications after performing an abstract, keyword, and full-text analysis.

**Theoretical Considerations.** Most publications ($n = 12$) have merely centered the theoretical relevance of cognitive load (e.g., Herm et al., 2021) and assumed that reduced cognitive load positively affects end-user performance (e.g., Hemmer et al., 2021) and assists end-users to solve the task faster (Bertrand et al., 2022). It is also hypothesized that increased problem complexity might be perceived as cognitively demanding (Cai et al., 2019). Similarly, research suggests that increased cognitive load reduces end-user trust in the system (e.g., Sultana and Nemati, 2021). In this context, research has derived tentative design principles (Fahse et al., 2022a) or design frameworks that assume reduced information granularity diminishes cognitive load (Barda et al., 2020).

**Empirical Research.** Only scarce research ($n = 5$) has focused on testing XAI's cognitive load. These contributions have mainly compared a single XAI implementation package or single XAI explanation type with a black-box implementation (e.g., Abdul et al., 2020) under simplified conditions, such as proxy tasks (Buçinca et al., 2020). From that, these contributions provide first evidence, that increased explainability will reduce end-user's cognitive load (Kulesza et al., 2013). In addition, research has focused on the connection between the end-user's cognitive load and their confidence or trust (Davis et al., 2020, Karran et al., 2022), implying that increased cognitive load slightly negatively affects perceived confidence and trust.

**Research Gap.** In summary, this sparse stream of research contains merely a handful of theoretical and empirical contributions. Regarding the former, theoretical considerations already hypothesize that use case complexity may affect end-users' cognitive load, which in turn affects task performance, task time, and trust. Concerning the latter, previous empirical contributions have mainly examined the cognitive load of end-users on a single XAI explanation package or type. Most strikingly, there is currently no research contribution that examines multiple implementation-dependent XAI explanation types simultaneously to provide conceptual guidance for a domain-independent XAI-based DSS application. Furthermore, while research has emphasized the potential impact of cognitive load on end-user task performance and time to solve a particular task, empirical evidence is lacking. As a result, to the best of our knowledge, we are the first to use these preliminary results to perform a holistic empirical cognitive load evaluation of these implementation-independent XAI explanation types and their impact on task performance and task time.

## 2.4 Measurement Model

To conduct our research, we derive and test hypotheses to investigate the cognitive load of the aforementioned *explanation types* and their effects on *task performance* and *task time*. Beyond this hypotheses testing, the findings are then used as input for the mental efficiency metric of Paas et al. (2016) to enable a summative evaluation (cf. Table 2). In the following, we describe the derivation of the hypotheses for our RQ and provide an overview of the measurement model.

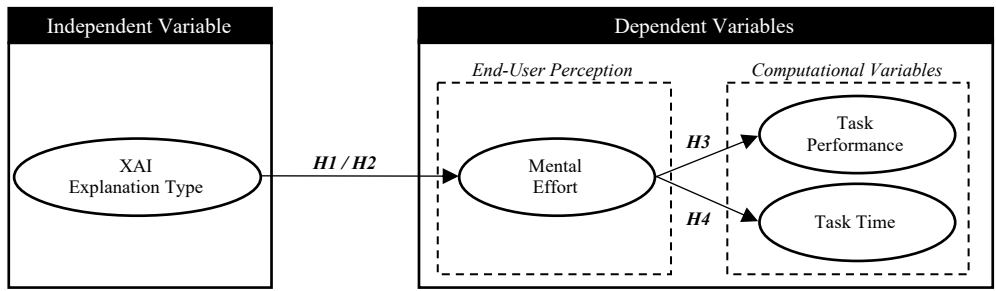

*Figure 1.*     *Measurement Model*

In line with the CFT, we derive a group structure consisting of one independent variable and three dependent variables. The independent variable is the type of *XAI explanation*, while the dependent variables are *mental effort, task performance,* and *task time*. Here, the independent variable represents the choice of *XAI explanation* types to provide reasoning for the DL model's decision logic. The dependent variable of *mental effort*, defined as the total sum of cognitive processing that a human is engaged in, indicates the perceived level of cognitive load required to comprehend the provided *XAI explanation* for task solving (Leppink and Pérez-Fuster, 2019, Paas and Van Merriënboer, 1993). Similarly, the dependent variable of *task performance* results from the end-user's ability to use the provided *XAI explanation* to solve a task within a use case. Finally, the dependent variable of *task time* results from the time required by an end-user to solve a task when using an *XAI explanation*.

First, we assume that assisting an end-user with any type of XAI explanation would reduce the mental effort required to comprehend an ML model's reasoning for a classification (Mohseni et al., 2021). This is because these explanations pinpoint towards relevant parts of the observation for the model's classification, compared with end-users who have to figure this out for themselves (Meske et al., 2022). Therefore, we propose the following hypothesis:

*H1: Any type of XAI explanation reduces mental effort compared with no explanation.*

Second, while research suggests that XAI explanations differ in terms of their perceived explainability (Herm et al., 2023), we assume that this degree of explainability is in line with the perceived *mental* *effort* required to comprehend the reasoning of an ML model. That is, while explanation types such as *Why* and *Why-Not* explanations are local explanations – and therefore have a more straightforward explanatory fashion and scope than global explanation types (e.g., *How*) (Buçinca et al., 2020, Speith, 2022) – we hypothesize that variations in explanatory scope and style would result in a different level of required mental effort for each *XAI explanation* type. Thus, we formulate the following hypothesis:

*H2: Each type of XAI explanation differs in terms of mental effort.*

Third, providing information that requires a high cognitive load may overwhelm people during task solving, resulting in weak *task performance* (Hemmer et al., 2021). This may be the case when too much information is presented in a complex scenario, wherein humans are either incapable of comprehending all of it or deliberating among the levels of relevance within it (Hudon et al., 2021). Bringing this into an XAI perspective, we hypothesize that *XAI explanation* types that require less mental effort would improve end-user *task performance*. Therefore, we propose the following hypothesis:

*H3: Less mental effort when using XAI explanations leads to improved end-user task performance.*

Fourth, in research, cognitive load is considered as the number of items processed within a limited time period (Leppink et al., 2014); thus, it impedes any other cognitive tasks or activities (Barrouillet et al., 2007). That is, tasks that require a relatively significant amount of time to solve are perceived as requiring increased *mental effort*, resulting in a linear relation (Leppink and Pérez-Fuster, 2019, Otto and Daw, 2019). Hence, we hypothesize that *XAI explanations* that require less *mental effort* would help end-users to solve tasks faster than explanations that demand more *mental effort*. Therefore, we formulate the following hypothesis:

*H4: Less mental effort when using XAI explanations leads to reduced end-user task time.*

## 3 Research Design

### 3.1 Methodology Overview

We follow the methodology of Müller et al. (2016) to ensure the rigor of our research. This involves a four-step process, namely *1)* RQ, *2)* data collection, *3)* data analysis, and *4)* results interpretation. Figure 2 presents an overview of the research design, followed by descriptions of the four research steps.

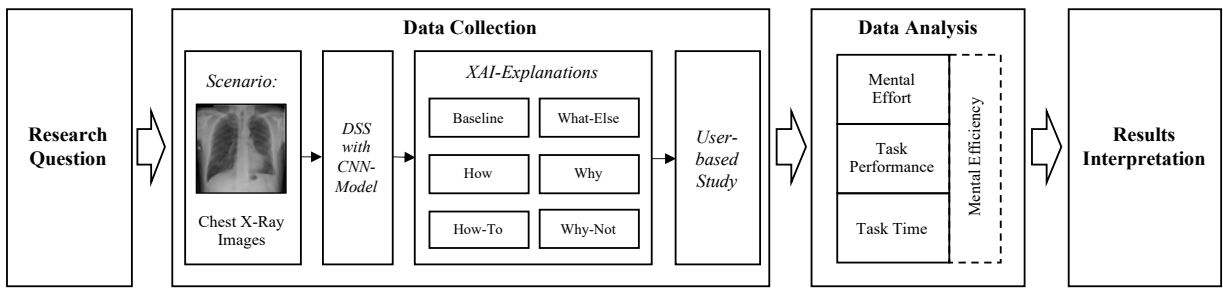

*Figure 2.     Overview of the Research Design*

*1) RQ*: Based on the structure literature review, we found that only a handful of contributions have investigated various XAI explanation types in a holistic and implementation-independent manner. Moreover, research assumes that these explanations differ in their cognitive load. Building on this knowledge gap, we derived hypotheses to investigate this assumption and further demonstrate whether this also affects task performance and task time. We use these findings to calculate the mental efficiency of these XAI explanation types. *2) Data collection*: We use a publicly available dataset of COVID-19 chest X-ray images, a DL model, and the aforementioned XAI explanation types to test our hypotheses through a user-based study. *3) Data analysis*: From the analysis, we derive a knowledge base for our research. *4) Results interpretation*: Ultimately, we answer our RQ and derive implications for research and practice.

## 3.2 Survey Design

To answer our RQ, we focused on a high-stakes use case from medicine. Specifically, we used chest X-ray images of COVID-19-infected and healthy humans (Tawsifur et al., 2022) to train a DL model – a convolutional neural network (CNN) – consisting of 11 layers, which yields a classification accuracy of 96.43% on the validation set. Then, we had the CNN classify several images of infected and healthy humans and enriched the images with the explanation style of the aforementioned XAI explanation types from Section 2.1.

Following the study design of Herm et al. (2023), we chose a within-subjects design for our study. First, we asked participants about their demographics, introduced the high-stakes use case, and described how the XAI-based DSS operates, enabling them to put themselves in the position of a physician deciding on a patient's well-being. Subsequently, we asked each participant to perform one assignment for every explanation type: Within each assignment, they received an input image of a chest X-ray, the corresponding XAI augmented image (XAI explanation), and a comprehensive description of the XAI explanation. For each explanation type, we designed two variants, one image with an infected chest and one for healthy patients. Only one variant is shown at a time (evenly and randomly distributed). Using the provided XAI explanation, each participant was asked to classify whether the depicted chest is infected with COVID-19 or not. Then, they were asked to rate the mental effort required for this classification task on a seven-point Likert scale (extremely low to extremely high).

Using their classification, we measured their (task) performance (correct or incorrect) and clocked the required time to complete the task (task time). Both measurements were performed for every assignment and every participant. An example of the study design for the explanation type *Why* is presented in Figure 3. See Herm (2023) for the complete questionnaire.

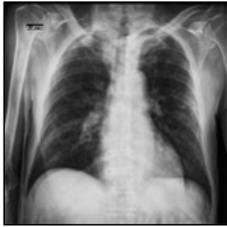

*Figure 3.         Example of the Study Design*

To avoid bias, we did not present the performance metrics of the used CNN (performance bias); did not use colors nor representations of XAI implementation packages (e.g., SHAP) to avoid confirmation bias; avoided learning effects through randomization; and only provided a comprehensive description for the XAI explanation to avoid forcing anchoring bias. Additionally, we incorporated several mechanisms, including attention checks, to ensure the validity of responses. Furthermore, we asked an XAI researcher to appraise our study design and a physician to review whether the classification tasks are equally difficult. Also, we conducted a preliminary study to test its validity. As we focused on the actual end-user of an XAI-based DSS, we targeted novice users in terms of AI experience. That is, we focused on prospective physicians currently enrolled as medical students, since experienced physicians might exhibit bias toward XAI-based DSS, and moreover, we wanted to focus on the future healthcare workforce (Herm et al., 2023, Logg et al., 2019).

## 4 Data Analysis

### 4.1 Survey Overview and Demographics

To recruit our participants, we used the *Prolific.co* platform, where we offered a monetary incentive of £10 per hour. Using this platform, we were able to specify and address our target group of prospective physicians (Peer et al., 2017). For this purpose, we gathered feedback from $n = 271$ participants. Since we performed several validation checks, such as randomly completed questionnaires, time-based outliers, lazy patterns, and control questions, we considered feedback from $n = 246$ participants to be optimal for our study. Among these, $n = 130$ participants were female, $n = 115$ were male, and $n = 1$ was diverse. Since we targeted enrolled students, $n = 12$ participants were younger than 20 years, $n = 193$ were 20–30 years old, and $n = 41$ were older than 31 years. They were located in Europe ($n = 125$), North America ($n = 64$), or Africa ($n = 50$). Regarding AI experience, $n = 95$ had no experience, while $n = 112$ had fewer than 2 years, and only $n = 39$ had more than 2 years. Further, $n = 84$ participants had less than 2 years of experience in medicine, $n = 101$ had 2–5 years, and $n = 61$ had more than 6 years.

### 4.2 Data Results

First, we provide an overview of the results and their distribution for the dependent variables for each explanation type (cf. Table 2). Subsequently, we utilize the findings to test our hypotheses in Table 3.

**Descriptive Statistics.** Table 2 highlights the results of the dependent variables: First, the mental effort findings, including medians and deviations, are plotted in Figure *a)*. Second, the total numbers of correct and incorrect answers are presented in Figure *b)*. Here, the average task performance is calculated by the ratio between correct and incorrect answers per type. Third, an overview of the distribution and kernel density of the time required per task is provided in Figure *c)*. These results are also summarized in tabular form. Thereon, we calculate the mental efficiency of the explanation types (Paas et al., 2016).

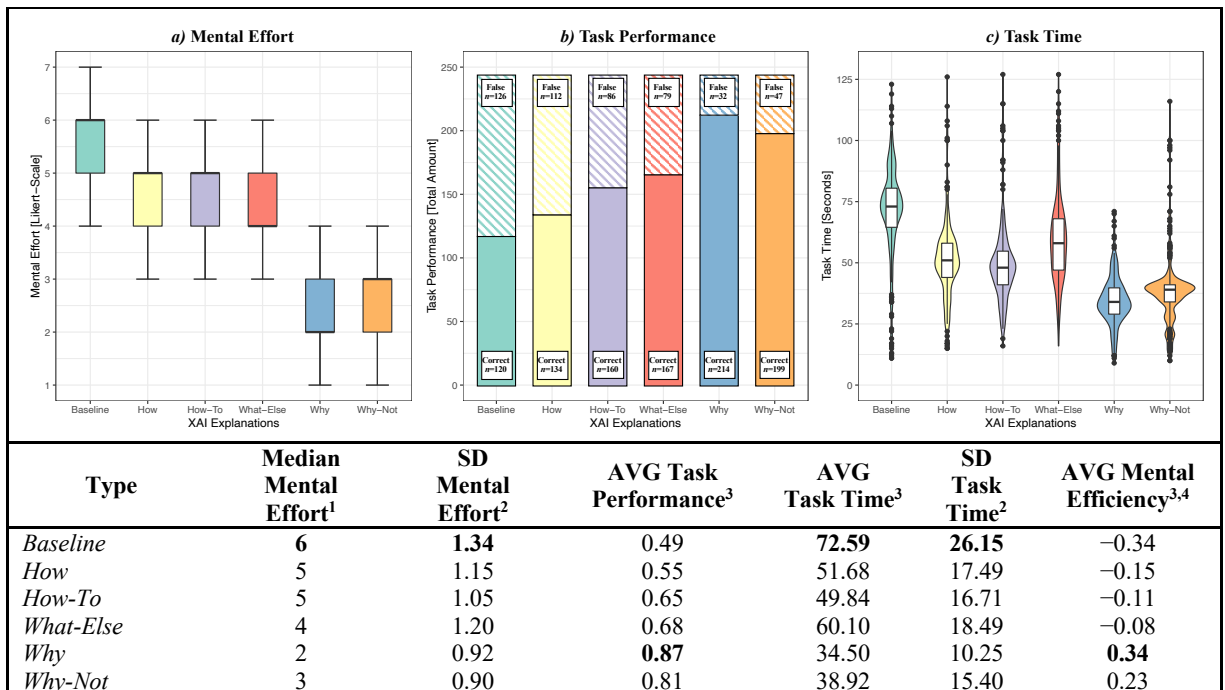

| Type | Median Mental Effort[1] | SD Mental Effort[2] | AVG Task Performance[3] | AVG Task Time[3] | SD Task Time[2] | AVG Mental Efficiency[3,4] |
|---|---|---|---|---|---|---|
| *Baseline* | **6** | **1.34** | 0.49 | **72.59** | **26.15** | −0.34 |
| *How* | 5 | 1.15 | 0.55 | 51.68 | 17.49 | −0.15 |
| *How-To* | 5 | 1.05 | 0.65 | 49.84 | 16.71 | −0.11 |
| *What-Else* | 4 | 1.20 | 0.68 | 60.10 | 18.49 | −0.08 |
| *Why* | 2 | 0.92 | **0.87** | 34.50 | 10.25 | **0.34** |
| *Why-Not* | 3 | 0.90 | 0.81 | 38.92 | 15.40 | 0.23 |

Legend: *1)* Median on a 7-point Likert scale [1,7] according to Boone and Boone (2012); participant-fixed model (LSDV): RSE: 1.047, multiple $R^2 = 0.632$, adjusted $R^2 = 0.563$, $F=7.741$, $p < 2.2e-16$; *2)* standard deviation of mental effort/ task time; *3)* average of task performance [0,1]/ task time (in seconds)/ mental efficiency {−1..1}; *4)* mental efficiency as $ME = \frac{Z_{task\,performance} \times Z_{task\,time} - Z_{mental\,effort}}{\sqrt{2}}$ adapted from Paas et al. (2016), mental effort and task performance standardized and task time standardized and reversed scale applied for computation.

Table 2.     *Descriptive Results of Cognitive Load Questionnaire*

*Mental Effort.* The absence of an XAI explanation (*Baseline*) led to the highest required mental effort in this study (median = 6). The local explanations *Why* (median = 2) and *Why-Not* (median = 3) required the least mental effort to solve the task. By contrast, the global explanation *How* (median = 5) and the *How-To* explanation (median = 5) required the most mental effort across all XAI explanation types. Within this range, providing multiple images for a task to indicate similar examples (*What-Else*, median = 4) was rated as requiring moderate mental effort.

*Task Performance.* Regarding task-solving performance, without any XAI explanation (*Baseline*), the participants solved approximately 49% of the tasks correctly. Consistent with the mental effort results, using the explanations *Why* (87%) and *Why-Not* (81%) led to the highest task performance. When participants were supplied with a global explanation (*How*), their task performance increased slightly (55%) compared with no XAI explanation. Finally, the explanation types *How-To* (65%) and *What-Else* (68%) were in the middle of this comparison.

*Task Time.* When participants did not use XAI explanations (*Baseline*), they took the longest time on average (72.59 sec) to solve a task. By contrast, the explanations *Why* (34.50 sec) and *Why-Not* (38.92 sec) almost halved the elapsed time. We noticed that these explanation types exhibited a high density around this meantime compared with explanations the *What-Else* or *How-To*. In this respect, the mean task times of the *How-To* (49.84 sec), *How* (51.68 sec), and *What-Else* (60.10 sec) explanations were much closer to the baseline than those of the *Why* and *Why-Not* explanations.

*Mental Efficiency.* Since an ME above null would indicate that the end-users' performance was higher than expected compared with the mental effort invested (Paas et al., 2016), the explanations *Why* (0.34) and *Why-Not* (0.23) can be considered highly efficient. By contrast, the most mental effort was required to solve a task when no XAI explanation (*Baseline*, −0.34) or *How* explanation (−0.15) was presented. The explanations *How-To* (−0.11) and *What-Else* (−0.08) also performed slightly better in this calculation but still yielded negative values.

**Hypotheses Testing.** To test our hypotheses (cf. Section 2.4), we follow Motulsky (2014) and apply different testing methods for H1–H4 depending on the type of test case, as demonstrated in Table 3. For each hypothesis, the results are plotted and then the statistical method, resulting *p*-value, and corresponding decision of acceptance or rejection are provided below.

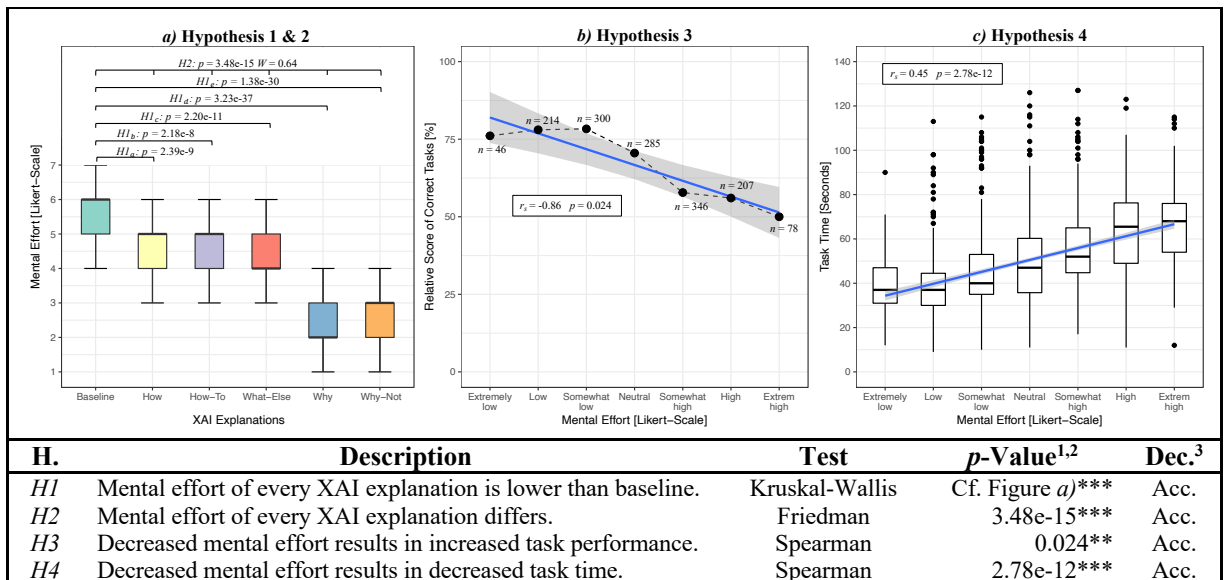

| H. | Description | Test | *p*-Value[1,2] | Dec.[3] |
|---|---|---|---|---|
| H1 | Mental effort of every XAI explanation is lower than baseline. | Kruskal-Wallis | Cf. Figure *a)*\*\*\* | Acc. |
| H2 | Mental effort of every XAI explanation differs. | Friedman | 3.48e-15\*\*\* | Acc. |
| H3 | Decreased mental effort results in increased task performance. | Spearman | 0.024\*\* | Acc. |
| H4 | Decreased mental effort results in decreased task time. | Spearman | 2.78e-12\*\*\* | Acc. |

Legend: *1)* \* <0.10, \*\* <0.05, \*\*\* <0.001; *2)* for H1, each test yielded high significance; *3)* decision of acceptance (acc.) or rejection (rej.) of the hypothesis.

*Table 3.        Results of Hypotheses Testing*

Using the results in Table 3, we decide whether to accept or reject our hypotheses as follows: First, to test whether providing an XAI explanation reduces the mental effort required to solve a task compared with no explanation (*Baseline*) (**H1**), we performed five Kruskal-Wallis tests that compared each XAI explanation with our baseline individually. This yielded highly significant results for each comparison, which confirm H1. Second, to investigate whether, due to the different explanatory scopes, each XAI explanation led to different levels of perceived mental effort, we performed a Friedman test and compared all types. Since this procedure revealed highly significant results, we accept **H2**. Third, to test whether using XAI explanations perceived as less demanding in terms of mental effort led to higher task performance (**H3**), we performed a Spearman correlation test to find an association between these two dependent variables. We found evidence of a significant correlation and thus accept H3. Finally, to test whether lower mental effort also correlates with lower task time (**H4**), we performed a Spearman correlation test to detect an association between mental effort and task time. We obtained a highly significant correlation, confirming H4.

# 5 Results Interpretation

## 5.1 Discussion of Results

To address our RQ, we interpret the results presented in Section 4.2 compromising end-users' cognitive load, task performance, and task time. Subsequently, we discuss the computed metric mental efficiency (cf. Table 2) to combine the findings of these dependent variables.

**Impact of XAI Explanation Types on Cognitive Load.** Recent research (Karran et al., 2022) already assumed that any type of XAI explanation assists the end-user in solving tasks, thereby reducing the required cognitive effort, as any type of explanation helps to render the end-user's mental model more congruent with the task problem compared with no explanation. We support this assumption through H1. Conversely, unstable explanations can influence the mental model. Thus, explanations are likely to impact end-user trust, especially when abductive reasoning is engaged (e.g., in complex or high-stake use cases) (Lakkaraju and Bastani, 2020). Hence, providing explanations to end-users encourages them to simplify their mental model based on the information supplied and potentially rely solely on the ML model's rationale, which could theoretically lead to mispredictions (Janssen et al., 2022).

As previous work (e.g., Buçinca et al., 2020) has already assumed that end-users perceive explanations to be individually demanding due to variations in the amount of information available and the style of explanation, we found that each XAI explanation type differs in terms of mental effort (H2). These results are reinforced, as end-users reasoning can be distinguished into a rational or an intuitive cognitive process, emphasizing a salience features evaluation or a systematic evaluation (Hamilton et al., 2016). In this regard, local explanations, namely the explanation types *Why* and *Why-Not,* yielded the lowest median and standard deviation among our results concerning mental effort. While this is consistent with Weerts et al. (2019)'s empirical study, which tested a local explanation using the SHAP package, Herm et al. (2021) also found that using this package can lead to misinterpretation due to confusing color palettes or additional information. Combining our research and related studies, we expected the use of color-free *Why* or *Why-Not* explanations to impose the least mental effort on end-users. By contrast, the XAI explanation *How* returned the highest mental effort score of our study and ranked close to the baseline. In research, this type of explanation is highly debated as it provides the most information compared with other types; hence, it can be presumed to have the highest explanatory scope (Hudon et al., 2021). Still, it could possibly also overwhelm non-ML experts (Fürnkranz et al., 2020).

Comparing our findings with Herm et al. (2023)'s explainability evaluation and their assumption that explainability is concomitant with cognitive load, we observe tendencies indicating a correlation between the two factors. That is, when comparing explainability and mental effort, comparative results emerged for the *Baseline*, *How*, *How-To*, and *Why* explanation types. In turn, we identified differences for *Why-Not* and *What-Else* explanation types. Here, end-users perceive *What-Else* explanations as more explainable (presumably) due to their information scope, but requiring increased mental effort to comprehend, which is congruent with the assumption of Miller (2019). Still, in this clinical context, distinct requirements for the explanatory scope and domain-specific regulations necessitate a detailed level of granularity (Ghanvatkar and Rajan, 2022). Also, contrary to the research of Herm et al. (2023), the local *How-To* explanation demanded an increased mental effort compared to the global *What-Else* explanation. Reflecting Sultana and Nemati (2021), we surmised that this was due to the complexity of our task, which might be different with fewer features or image segments. Given the broad distribution of task time when using the *What-Else* explanation, we assumed that mental effort also depends on whether end-users grasp or struggle with this type of explanation. Still, researchers argued that this type is relatively facile to realize and its application merits prior training of end-users (Kim et al., 2016). Lastly, considering local explanation types (*Why*, *Why-Not*) perceived best, these explanation types might cause difficulties, as end-users tend to rely on features that are highlighted by the explanations (Bauer et al., 2023). Hence, guidelines are required to ensure the application of XAI in high-stakes use cases (Kloker et al., 2022).

**Impact of Cognitive Load on Task Performance.** As we obtained significant results for a linear correlation between perceived mental effort and task performance (H3), this denotes a general surplus in end-user task performance. Yet, our results are consistent with Fahse et al. (2022b)'s and Hemmer et al. (2021)'s assumptions that cognitive load and task performance are diametrically related. However, we recognize some relative outliers. In particular, one might expect the responses rated "extremely low" in terms of mental effort to have produced the best task performance results; however, we found moderate to high relative task performance. This could be due to the relatively small sample size, and an outlier could skew the results. In addition, a related study already found that participants become negligent when a task is not mentally demanding (so-called "cognitive underload"), and thus, errors accumulate (Lavie, 2010). Nevertheless, when explanations require less cognitive load, the end-user's mental model is more capable of retrieving information and recognizing new circumstances more quickly (Abdul et al., 2020). In this regard, these results should be taken with a grain of salt as we targeted novice medical end-users who were unlikely to have actively used an XAI-based DSS before. These results may change once end-users are taught how to use these types of systems or use them more frequently due to the iterative learning process (Engström et al., 2017). Thus, the explanations *What-Else* or *How* may be favored due to their increased information scope but cease to overwhelm eligible participants.

**Impact of Cognitive Load on Task Time.** Given that research has previously assumed a linear relationship between perceived mental effort and task time (Bertrand et al., 2022, Leppink and Pérez-Fuster, 2019), highly significant results also emerged for H4. Here, the task time per level of mental effort was consistent with the results and the corresponding mental effort medians. The high density within the *Why* and *Why-Not* explanation types indicates general straightforward intelligibility for novice end-users. Surprisingly, considering this linear relationship, the global *What-Else* explanation was perceived as less demanding, yet participants were able to solve our tasks faster with the local *How-To* explanation. We attribute this to the nature of the explanation, as participants might not have used such support before. Further, Buçinca et al. (2020) argue that increased task time results from end-users' commitment to comprehend the provided explanation, as they may not trust the AI's recommendations. Conversely, a comparatively low task time could indicate over-reliance on explanations. In research, the task time factor is highly controversial: Liao and Varshney (2022) stated that in the absence of time pressure, more complex explanations should be preferred as an end-user is able to iteratively discover new relationships within the explanations. Contrary, in real-world applications, a thorough evaluation process is temporarily infeasible (Shaft and Vessey, 2006). However, research has already discussed that increased task time may impact end-user satisfaction (Hsiao et al., 2021).

**Mental Efficiency Ramifications.** Local explanation types perform best regarding mental efficiency, resulting in a positive value. Therefore, end-users employing *Why* and *Why-Not* explanations exceed the performance-mental effort ratio, leading to a higher-than-expected result (Paas et al., 2016). However, compared to the cognitive load results of *What-Else* explanations, these explanations are about the participants' expected value. That is, while our findings imply relatively high mental effort, this mental efficiency result hints at relatively high end-user commitment levels, which could be consistent with the perceived explainability results of Herm et al. (2023). Comparatively, there is limited research evaluating XAI explanation metrics that incorporate end-user understanding (Gentile et al., 2021). Merely Ghanvatkar and Rajan (2022) and Fahse et al. (2022b) derived metrics to measure a person's effectiveness. Since we target cognitive load, we also transfer the cognitive load into the context of XAI. Accordingly, we distinguish as follows: Ghanvatkar and Rajan (2022) consider layer-wise relevance propagation (global explanation) to be the most effective as it provides the utmost information. However, this can also be critical as end-users may be unable to complete a task when overloaded. Conversely, for domain-specific requirements (e.g., in a clinical context), XAI explanations mandate a certain level of information, raising the importance to focus on effectiveness rather than efficiency. With this in mind, while one should not rely solely on effectiveness or efficiency, the trade-off should be determined based on the use case at hand (Forsythe et al., 2014).

## 5.2 Implications, Limitations, and Future Research

**Theoretical Implications.** While research (e.g., Buçinca et al., 2020, Hudon et al., 2021) has only partly investigated the cognitive load of singular XAI implementations, holistic comparisons of distinct implementation-independent XAI explanation types are lacking. This is especially critical when considering potential bias, which may confuse end-users or even force them to make erroneous decisions (Nourani et al., 2022). From a theoretical perspective, we have contributed to the existing body of human-technology interaction knowledge, one of the cores in IS research (Riefle and Benz, 2021), by researching XAI's cognitive load and related effects on task performance and task time to ultimately derive a mental efficiency metric for the evaluation of XAI explanations. To best of our knowledge, we are the first to place this type of metric into the context of XAI and thus also take the end-user's mental model into account. Likewise, by directly comparing task performance and task time to cognitive load, we contribute to this relatively sparse body of knowledge in XAI research.

Here, we demonstrate that XAI explanations are essential for recommendation-based decision support because they reduce cognitive load, increase task performance, and reduce task time. Consequently, local explanations perform best in terms of mental efficiency. Following the ongoing (IS research) debate on the selection of explanation types (Gregor and Benbasat, 1999, Herm et al., 2023, Meske et al., 2022), we therefore provide initial insights on cognitive load for implementation-independent XAI explanation types. Drawing on this, this jigsaw piece contributes to the overall puzzle of the end-user's heuristic mental model. Although we did not measure trust and reliance during the experiment, we can identify some tendencies that could indicate end-user over-reliance especially on more straightforward explanations (e.g., *Why*). Thus, while Miller (2019) posits four requirements for the goodness of an explanation, our research indicates that focusing on causal reasoning and selective representation likely facilitates misclassification when the AI's recommendation is inaccurate. This may also be related to the type of end-users, as we focus on young professionals who tend to use the XAI-based DSS for support and pattern learning. In contrast, experts are prone to focus on using these explanations for verification (Gregor and Benbasat, 1999) and AI-experienced individuals have more reservations about AI explanations (Herm et al., 2022). In this regard, our results may differ as we focus on additional end-user groups, which means that the role of explanations may vary (Bauer et al., 2023). As recent research has shown that providing an explanation has a positive impact on trust and attitudes toward an AI-enabled DSS (Wanner et al., 2022), this end-user over-reliance can lead to unwarranted trust that results in automation bias, even in high-risk use cases (Jacovi et al., 2021). Therefore, cognitive forcing strategies should accompany the utilization of XAI-based explanations.

Although recent IS research calls for a paradigm shift in XAI, proposing the application of hypothesis-driven support instead of recommendation-driven support to accommodate the end-user's cognitive process (Miller, 2023); the evaluation of explanations remains critical to ensure appropriate knowledge transfer of inferred evidence for an end-user action. Moreover, this approach forces end-users to be more committed, which increases their cognitive load and consequently emphasizes the need for mentally efficient explanations. To this end, we further advance the theoretical debate through the provision of a sociotechnical metric to evaluate XAI explanation types. As our results can be considered as a cognitive load-centered starting point for the discussion on the role of explanations in IS research, it currently lacks longitudinal analysis to determine additional aspects such as learning effects. This includes a combined study of other factors such as trust, acceptance, and satisfaction, which appear to be essential to understand the end-users' heuristic behavior. Ultimately, the benefits of providing XAI-based explanations in DSS will facilitate the integration of ML algorithms into organizational information systems, thus embedding the potentials of AI into intelligent systems (Gregor and Benbasat, 1999, Wanner et al., 2022).

**Practical Implications.** In the early days of XAI research, XAI was seen as the silver bullet for end-user adoption of AI in any use case (Goebel et al., 2018); however, we found significant differences in perceived cognitive load, task performance, and required task time among the XAI explanation types. Thus, several considerations must be made: First, our research identified that developers of recent XAI implementations (cf. Table 1) must reconsider their applications, building upon our results, with respect to sociotechnical factors (e.g., cognitive load) and redesign them to match end-users' mental model. Second, we demonstrated that not every explanation type is appropriate for every situation; thus, practitioners must determine an appropriate explanation based on various factors, such as performance constraints, time constraints, or use case requirements. Third, it should be considered that explanations are usually a simplification of the ML model's rationale, and therefore, they are unlikely to contain the entire decision logic, which may include bias, adversarial attacks, or open Pandora's box due to the non-applicability of the GDPR (Slack et al., 2021). That is, relying on inherently explainable ML models could be essential once a defined performance threshold is fulfilled (Rudin, 2019).

**Limitations and Future Research.** Like any empirical study, ours has its limitations. We focused our fundamental XAI research on implementation-independent explanation types to derive unbiased insights for further XAI development, which must be translated into concrete and use case-specific applications. Specifically, researchers and practitioners must integrate these insights into their XAI explanations and then re-evaluate their improved artifacts. Our research could also be expanded as follows: Following the triangulation approach of Tams et al. (2014), future research should validate our findings by using further complementary measurement approaches such as eye-tracking studies and electro-encephalograms to identify how end-users behave when using these explanation types. Second, our study should be expanded by examining additional factors to determine a holistic understanding of an end-user's behavior. Third, given the assumption of a tendency toward over-reliance within our study, future research should carry out dedicated research to examine potential trust miscalibrations. This includes investigating whether end-users are able to detect erroneous recommendations from the XAI-based DSS. Fourth, while we focused on a representative use case from the medical field, future research should leverage our findings to conduct further studies in other high-stakes use cases and with different types of end-user groups. However, our results lay the foundation for the end-user-centered design of XAI explanations and the derivation of design principles for XAI-based DSSs (Herm et al., 2022).

# 6 Conclusion

AI is emerging as a frontier of computational advances for mimicking or surpassing human intelligence. However, in high-stake decision-making use cases, the models' internal decision logic hinders the use of DL-based applications due to being incomprehensible to end-users and thus reducing their willingness (Berente et al., 2021, Wanner et al., 2022). XAI has gained momentum by making these black-boxes understandable while maintaining the predictive power of the underlying model (Janiesch et al., 2021). Despite the proliferation of XAI applications, actual end-users are not sufficiently addressed (van der Waa et al., 2021). Unsurprisingly, using these systems for high-stakes use cases will likely result in overwhelmed and stressed end-users, which might not perform well due to high cognitive load (Hudon et al., 2021). In this regard, actual user-centered XAI research is relatively limited (Laato et al., 2022). To address this knowledge gap on end-users' cognitive behavior, we used COVID-19 X-ray images to conduct an empirical study, thereby investigating how distinct implementation-independent XAI explanation types affect end-users' cognitive load, task performance, and the time required to solve a task. Combining our results, we calculate the mental efficiency of these explanation types. This facilitates an in-depth empirical study and thus, the derivation of implications for future research and practice. In doing so, we contributed to the current body of XAI knowledge to surmount the *"inmates running the asylum"* situation (Miller, 2019) in sociotechnical XAI research.